\documentclass[letterpaper]{article} 
\usepackage{aaai24}  
\usepackage{times}  
\usepackage{helvet}  
\usepackage{courier}  
\usepackage[hyphens]{url}  
\usepackage{graphicx} 
\usepackage{natbib}  
\usepackage{caption} 
\usepackage{algorithm}
\usepackage{algpseudocode}
\usepackage{newfloat}
\usepackage{listings}
\DeclareCaptionStyle{ruled}{labelfont=normalfont,labelsep=colon,strut=off} 
\floatstyle{ruled}
\newfloat{listing}{tb}{lst}{}
\floatname{listing}{Listing}
\usepackage{amsthm}
\usepackage{thmtools} 
\usepackage{thm-restate}
\usepackage{amssymb}
\usepackage{amsmath}
\usepackage{algorithm}
\usepackage{algpseudocode}
\usepackage{booktabs}
\usepackage{array}

\title{Adversarially Balanced Representation for Continuous Treatment Effect Estimation}
\author{
    Amirreza Kazemi\textsuperscript{\rm 1}, Martin Ester\textsuperscript{\rm 1}
}
\affiliations{
    \textsuperscript{\rm 1}Department of Computing Science, Simon Fraser University\\
    \{aka208, ester\}@sfu.ca

}

\usepackage{bibentry}

\begin{document}

\maketitle

\begin{abstract}
Individual treatment effect (ITE) estimation requires adjusting for the covariate shift between populations with different treatments, and deep representation learning has shown great promise in learning a balanced representation of covariates. However the existing methods mostly consider the scenario of binary treatments. In this paper, we consider the more practical and challenging scenario in which the treatment is a continuous variable (e.g. dosage of a medication), and we address the two main challenges of this setup. We propose the adversarial counterfactual regression network (ACFR) that adversarially minimizes the representation imbalance in terms of KL divergence, and also maintains the impact of the treatment value on the outcome prediction by leveraging an attention mechanism. 
Theoretically we demonstrate that ACFR objective function is grounded in an upper bound on counterfactual outcome prediction error. 
Our experimental evaluation on semi-synthetic datasets demonstrates the empirical superiority of ACFR over a range of state-of-the-art methods.
\end{abstract}

\section{Introduction}
Estimating the individual treatment effect (ITE) from observational datasets has 
important applications in domains such as personalized medicine \cite{kent2018personalized, prosperi2020causal}, economics \cite{doi:10.1073/pnas.1510479113}, and recommendation systems \cite{10.1145/3383313.3412225}. The observational dataset contains units with their covariates $X$, the assigned treatment $T$, and outcome after intervention $Y$  (also known as the factual outcome). Since the treatment assignment policy is unknown, there typically exists an inherent treatment-selection bias stemming from confounding covariates that influence both the treatment assignment and the outcome. Consequently, a causal ITE estimator requires adjusting for the covariate shift among different treatment populations \cite{bareinboim2022recovering, bareinboim2012controlling}.
\newline
\newline
In recent years, the representation learning approach \cite{pmlr-v48-johansson16, shalit2017estimating} has demonstrated remarkable success in adjusting for covariate shift. Briefly, the idea is to learn a balanced representation of covariates (rather than balancing the covariates themselves) using an encoder and then predicting the outcomes from the representation using an outcome prediction network.
However, the majority of the representation learning methods have been proposed for binary treatments \cite{hassanpour2019learning, zhang2021treatment, Yao2018RepresentationLF}, and incorporating continuous treatments into their architectures is challenging \cite{chu2023causal}. We elaborate on the associated challenges with continuous treatments in the following.
\newline
\newline
\textit{i) Balancing representation for continuous treatments}\\
To achieve a balanced representation $Z$, most existing methods minimize the shift in terms of Integral Probability Metric (IPM) distance \footnote{$
IPM_{G}(p, q) = \sup_{g \in G} \int_{S} g(s) \big(p(s) - q(s) \big) ds
$, where specifying $G$ leads to different distributional distance. } \cite{sriperumbudur2009integral} between the distributions of $P(Z, T)$ and $P(Z)\,P(T)$ as it allows bounding the counterfactual prediction error \cite{shalit2017estimating}.
Given that the marginal distribution $P(Z)$ is unknown, \cite{bellot2022generalization} proposed minimizing the IPM distance between $P(Z, T=t)$ and $P(Z, T=\lnot t)$ where $t$ denotes a treatment value in the data and $\lnot t$ denotes all treatments except $t$. Similarly, \cite{NEURIPS2022_390bb66a} suggested discretizing the treatment range into intervals and minimized the maximum IPM distance between the distributions of the two intervals. 
Despite minimizing an upper bound, these methods involve non-parametric approximations of several IPM distances in practice, which may be inaccurate for high-dimensional representation and small training data \cite{liang2019estimating}.
Furthermore, IPMs are by definition worst-case distances and and obtaining a treatment-invariant representation through IPM might be overly restrictive, potentially excluding important confounding factors for outcome prediction \cite{zhang2020learning}. 

There are also non-IPM representation learning methods. 
For instance, \cite{Du_2021,NEURIPS2020_e7c573c1} adopt an adversarial discriminator in order to balance the distributions in the latent representation, \cite{Yao2018RepresentationLF} proposed preserving the local similarity in the representation space, \cite{hassanpour2019learning, wu2022learning} learn a disentangled representation to distinguish different latent factors, \cite{zhang2020learning} enforce invertibility of the encoder function to prevent the loss of covariate information, and \cite{alaa2017bayesian} introduced a regularization scheme in order to generalize to counterfactual outcomes.
However, these methods lack theoretical justification \cite{Du_2021, hassanpour2019learning, Yao2018RepresentationLF} or their guarantees are limited to binary treatments \cite{zhang2020learning, alaa2017bayesian}.
\newline
\newline
\textit{ii) Treatment impact in outcome prediction network}\\
In order to predict outcomes of different treatments, the outcome prediction network on top of the representation needs to incorporate the treatment value. Considering the treatment variable as an input of the outcome prediction network along with the representation
causes overfitting the much higher-dimensional representation and largely limits the impact of the treatment value. Also, a distinct prediction head for each treatment value (as in the case of binary treatments) is not practical.  
Instead, \cite{Schwab_2020} proposed dose response network (DRNet) which divides the treatment range into intervals and consider a distinct network for the prediction of each interval. \cite{nie2021vcnet} proposed a varying coefficient network (VCNet) that involves the treatment value in the network parameters through spline functions of treatment. 
Nonetheless, both networks can not capture the dependency between representation and treatment effectively since the choice of spline functions in VCNet (or the intervals in DRNet) are made irrespective to representation values \cite{zhang2022exploring}.
\newline
\newline
In this paper,  we propose a representation learning method to accurately predict potential outcomes of a continuous treatment. We address the above challenges through the following contributions: 
\begin{enumerate}
  \item  We prove that under certain assumptions the counterfactual error is bounded by the factual error and the KL divergence between $P(Z)P(T)$ and $P(Z, T)$. Unlike the IPM distance, the KL divergence can be estimated parametrically, leading to a more reliable bound.
  \item 
  We propose Adversarial Balanced Counterfactual Regression (ACFR) network. ACFR minimizes the KL divergence using an adversarial game extracting a balanced representation for continuous treatments. ACFR also minimizes the factual prediction error by a cross-attention network that captures the complex dependency between treatment and the representation.
  \item 
  We conduct an experimental comparison of ACFR against state-of-the-art methods on semi-synthetic datasets, News and TCGA, and analyze the robustness to varying-levels of treatment-selection bias for the methods.
\end{enumerate}
\section{Problem Setup}
We assume a dataset of the form $D=\{x_i, t_i, y_i\}_{i=1}^N$, where $x_i \in 
\mathcal{X} \subseteq \mathbb{R}^d$ denotes the covariates of the $i$th unit, 
$t_i \in [0, 1]$ is the continuous treatment that unit $i$ received, and $y_i 
\in \mathcal{Y} \subseteq \mathbb{R}$ denotes the outcome of interest for unit 
$i$ after receiving treatment $t_i$. $N$ is the total number of units, and $d$ 
is the dimension size of covariates. We are interested in learning a machine 
learning model to predict the causal quantity $\mu(x, t) = E_{\mathcal{Y}}[Y(t) 
| X = x]$, which is the potential expected outcome under treatment $t$ for the 
individual with covariates $x$. Note that, unlike binary ITE, the goal is to 
predict all potential outcomes, not just the difference between them. Similar to 
previous works, we rely on the following standard assumptions to make treatment 
effects identifiable from an observational dataset.
\newline
\newline
\begin{restatable*}[ \textbf{Assumption 1 - Unconfoundedness}]{thm}{}
\label{ass1}
($\{Y_t\}_{t \in T} \perp T | X$). In words, given covariates, treatment and potential outcomes are independent.
\end{restatable*}
\newline
\newline
\begin{restatable*}[ \textbf{Assumption 2 - Overlap}]{thm}{} 
\label{ass2}
($P(T=t|X= x) > 0, \forall t \in [0,1], \forall x \in X$). In words, every unit receives treatment level $t$ with a probability greater than zero.
\end{restatable*}
\newline
\newline
With these assumptions, $\mu(x, t)$ can be rewritten as follows, and we can estimate it:
\[
\mu(x, t) = E_{\mathcal{Y}} [Y(t) | X= x] = E_{\mathcal{Y}} [Y | X = x, T = t]
\]

\section{Theoretical Analysis}
We analyze the properties of encoder function $\phi: \mathcal{X} \rightarrow \mathcal{Z}$, where $\mathcal{Z}$ is the representation space, outcome prediction function $h: \mathcal{Z} \times  [0, 1] \rightarrow \mathcal{Y}$ and loss function $L: \mathcal{Y} \times \mathcal{Y} \rightarrow \mathbb{R}^+$. 
\newline
\newline
\begin{restatable*}[\textbf{Definition 1}]{thm}{} 
\label{def1}
Define $\ell_{L, h, \phi}(x, t) = L(h(\phi(x), t), y): \mathcal{X} \times [0, 1] \rightarrow \mathbb{R}^+$ to be the unit-loss for a unit with covariate $x$ that is intervened with treatment $t$. Unit-loss $\ell_{L, h, \phi}(x, t)$ measures loss $L$ between the predicted outcome $\hat{y} = h(\phi(x), t)$ and the ground-truth outcome $y = \mu(x, t)$.
\end{restatable*}
\newline
\newline
Using the definition of unit-loss, we can define the expected prediction error of some treatment $t$ by marginalizing over the covariate distribution. As a result of treatment-selection bias, covariate distribution of samples having received treatment $t$ (factual) and samples not having received treatment $t$ (counterfactual) are different. We define factual error $\varepsilon_f^{\ell}(t)$ by marginalizing over $p(x|t)$ and counterfactual error $\varepsilon_{cf}^{\ell}(t)$ by marginalizing over $p(x)$ as follows.  
\begin{align*}
    &\varepsilon_f^{\ell}(t) = \int_{\mathcal{X}} \ell_{L, h, \phi}(x, t)\, p(x|t)\, dx \\
    &\varepsilon_{cf}^{\ell}(t) = \int_{\mathcal{T}' = [0, 1] - \{t\}}\int_{\mathcal{X}} \ell_{L, h, \phi}(x, t) p(x|t')\, dx \, dt' \\ &\qquad = \int_{\mathcal{X}} \ell_{L, h, \phi}(x, t)\, p(x)\, dx 
\end{align*}
We also define the expected error of all treatments by marginalizing over their range $[0, 1]$ as follows:
$\varepsilon_f^{\ell} = \int_{[0, 1]} \int_{\mathcal{X}} \ell_{L, h, \phi}(x, t)\, p(x, t) \, dx \, dt$ and $\varepsilon_{cf}^{\ell} = \int_{[0, 1]} \int_{\mathcal{X}} \ell_{L, h, \phi}(x, t)\, p(x)\,p(t)\,dx\, dt$.
Note that the expected factual error $\varepsilon_f^{\ell}$ integrates over joint distribution $p(x, t)$, and expected counterfactual error $\varepsilon_{cf}^{\ell}$ integrates over $p(x)\,p(t)$. We aim to reduce the distributional distance in representation space $Z$ to ensure that minimizing $\varepsilon_f^{\ell}$ results in minimizing $\varepsilon_{cf}^{\ell}$. We need the following assumption on encoder $\phi$ to ensure balancing properties generalize from the representation space to the covariate space.
\newline
\newline
\begin{restatable*}[\textbf{Assumption 3}]{thm}{ass3} 
\label{thm:ass3}
The encoder function $\phi$ is a twice-differentiable one-to-one mapping and the representation space $\mathcal{Z}$ is the image of $\mathcal{X}$ under $\phi$ with the induced distribution $p_{\phi}(z)$.
\end{restatable*}
\newline
\newline
We also need the following assumption to ensure the unit-loss is not arbitrary large for any $(x, t)$ pair. Constraint on the unit-loss function is required for the IPM distance specification as well.
\newline
\newline
\begin{restatable*}[\textbf{Assumption 4}]{thm}{ass4} 
Let $G$ be a class of functions with infinity norm less than 1, $G = \{g : \mathcal{Z} \times [0, 1] \rightarrow \mathbb{R}^+ \, | \,  ||g||_{\infty} \le 1 \}$. Then, there exist a constant $C > 0$ such that $
\frac{
    \ell_{L, h, \phi}(x, t)
}{
C
} \in G
$. This means for any $(x, t)$ we have $\frac{
    \ell_{L, h, \phi}(x, t)
}{
C
} \le 1$.
\end{restatable*}
\newline
\newline
Note that Assumptions 3 and 4 are common in representation learning literature \cite{shalit2017estimating, bellot2022generalization, hassanpour2019learning}.
Now we present our main theoretical results which demonstrates a bound on the expected counterfactual error $\varepsilon_{cf}^{\ell}$ consisting of the expected factual error $\varepsilon_f^{\ell}$ and the KL divergence between distributions $p_{\phi}(z, t)$ and $p_{\phi}(z)\,p(t)$.
\newline
\newline
\begin{restatable*}[\textbf{Proposition 1 - Counterfactual Generalization Bound}]{thm}{prop1} 
\textit{
Given the one-to-one encoder function $\phi : \mathcal{X} \rightarrow \mathcal{Z}$, the outcome prediction function $h : \mathcal{Z} \times [0, 1] \rightarrow \mathcal{Y}$, and the unit-loss function $\ell_{L, h, \phi}(x, t)$ that satisfies Assumption 4,
\[
\varepsilon_{cf}^{\ell} \le \varepsilon_f^{\ell} + C \sqrt{ 2\, D_{KL}\bigg(p_{\phi}(z, t)\,  \big|\big| \, p_{\phi}(z)p(t)\bigg)}
\]
}
\end{restatable*}
\\
Note that the KL divergence is non-negative and becomes zero if and only if two distributions are the same. Therefore, $D_{KL}\bigg(p_{\phi}(z, t) || p_{\phi}(z)p(t)\bigg) = 0$ implies $p_{\phi}(z, t)  = p_{\phi}(z)\,p(t)$ and $\varepsilon_f^\ell = \varepsilon_{cf}^\ell$.
We can also interpret the above bound from information theory perspective. Briefly, minimizing RHS of Proposition 1 results in representation $z$ which has maximum mutual information with outcome $y$ given treatment $t$ (equivalent to minimizing $\varepsilon_f^\ell$) and minimum mutual information with treatment $t$ (equivalent to minimizing KL divergence term).
\newline
\newline
For some applications, one might be interested in the treatment effect between two different treatments rather than predicting all counterfactual outcomes. For instance, in binary treatment setting it is standard to report the model performance in terms of precision of estimating heterogeneous effect (PEHE) \cite{Hill2011BayesianNM} which measures the squared difference between ground-truth treatment effect $\tau(x) = \mu(x, 1) - \mu(x, 0)$ and predicted treatment effect $\hat{\tau}(x) = h(\phi(x), 1) - h(\phi(x), 0)$. We define the continuous counterpart $\varepsilon_{pehe}(t_1, t_2)$ between two treatment levels $t_1$ and $t_2$ and present an upper bound on it in Proposition 2.
\newline
\newline
\begin{restatable*}[\textbf{Definition 2}]{thm}{def2} 
Define $\varepsilon_{pehe}(t_1, t_2) = \int_{\mathcal{X}} \big[ \big( \mu(x, t_1) - \mu(x, t_2) \big) - \big( h(\phi(x), t_1) - h(\phi(x), t_2) \big) \big]^2 p(x)dx$ to be the expected precision of estimating heterogeneous effect between treatment levels $t_1$ and $t_2$.
\end{restatable*}
\newline
\newline
\begin{restatable*}[\textbf{Proposition 2 - Precision of Estimating Heterogeneous Effect Bound}]{thm}{prop2}
\textit{Given the one-to-one encoder function $\phi$ and outcome prediction function $h$ as in Proposition 1, and a unit-loss function $\ell_{L, h, \phi}(x, t)$ that satisfies Assumption 4 and its associated $L$ is squared error $||.||^2$,
\small
\begin{align*}
   &\varepsilon_{pehe}(t_1, t_2) \le \varepsilon_f^{\ell}(t_1) \, + \, 
\varepsilon_f^{\ell}(t_2) \,  + \, \\
& C \Bigg[\sqrt{2\, D_{KL}\bigg(p_{\phi}(z) \, \big|\big| \, p_{\phi}(z|t_1)\bigg)} +
\sqrt{2\, D_{KL}\bigg(p_{\phi}(z) || p_{\phi}(z|t_2)\bigg)} \, \Bigg]
\end{align*}
}
\end{restatable*}
You can find the proof of Proposition 1 and 2 in the Appendix A.

\section{Method}
\begin{figure}[t]
    \centering
    \includegraphics[width=0.5\textwidth, height=0.32\textheight]{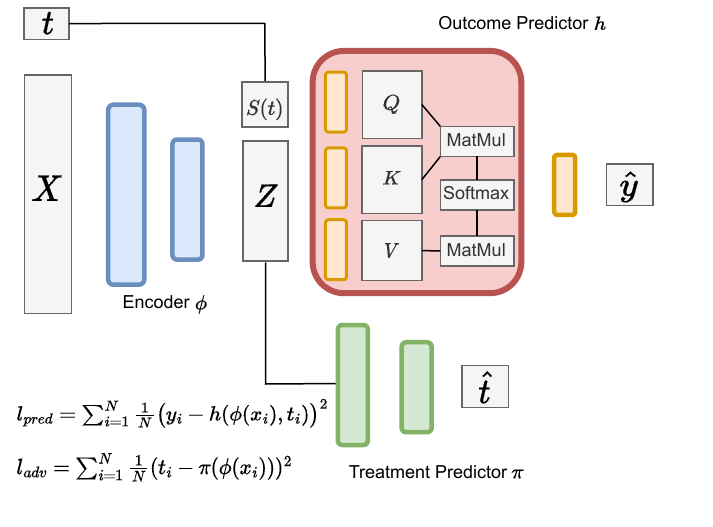}\caption{\textbf{The architecture of Adversarial CoutnerFactual Regression Network} consisting of three sub-networks encoder $\phi$, outcome predictor $h$, and treatment predictor $\pi$. Networks $\phi$ and $h$ are trained to minimize the outcome prediction loss $l_{pred}$, and networks $\phi$ and $\pi$ are trained to maximize / minimize adversarial loss $l_{adv}$. The encoder and treatment predictor are implemented using linear layers, and the outcome predictor network consists of a cross-attention module followed by a linear layer.}\label{fig:arch}
\end{figure}
Based on Proposition 1, we can control the counterfactual error by learning an encoder function $\phi$ and an outcome prediction function $h$ that jointly minimize distribution shift and factual outcome error. We propose our method that implements functions $\phi$ and $h$ using neural networks and is trained with an objective function inspired by Proposition 1. 
Figure \ref{fig:arch} illustrates the architecture of the ACFR network, consisting of an encoder network $\phi$, an outcome prediction network $h$, and a treatment prediction network $\pi$. The key aspects of ACFR, distribution shift minimization and minimization of outcome prediction error, are presented as follows.
\footnote{Note that in the previous section we assumed $\phi$ is a one-to-one mapping, however, we obtained better results empirically for all methods using a neural network as encoder.}
\begin{figure*}[t] 
    \centering
    \includegraphics[width=\textwidth]{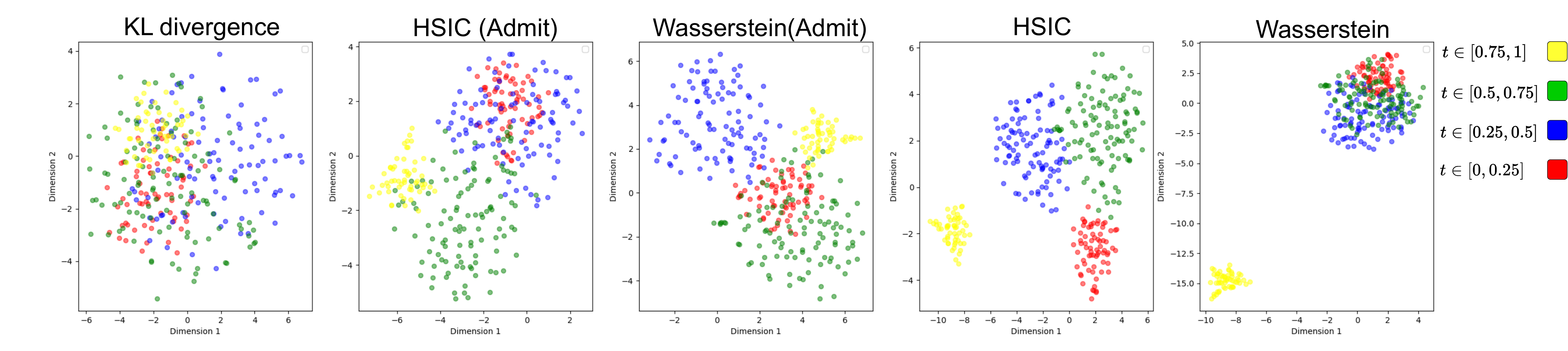} 
    \caption{\textbf{Tsne plot of latent representation $Z$ learned using different distributional distances}. After training each method on News dataset, we mapped validation samples into latent representation and plotted them using 2d tsne. We categorized the samples into 4 intervals with respect to their assigned treatment value and each interval corresponds to a color. We consider two important classes of IPM metrics, HSIC and Wasserstein. The treatment value is less distinguishable in the KL divergence representation followed by IPM-ADMIT (minimization with the algorithm proposed in \cite{wang2022generalization}), and IPM (minimization with the procedure proposed in \cite{bellot2022generalization})}
    \label{fig:dist}
\end{figure*}
\subsection{Distribution Shift Minimization}
As discussed earlier, in order to minimize distribution shift we aim to minimize the KL divergence term with respect to encoder $\phi$. The KL divergence can be rewritten as follows: 
\begin{align*}
   D_{KL}\bigg(p_{\phi}(z, t) \, || \, p_{\phi}(z)\,p(t)\bigg) &= I(T, Z; \phi) \\
&=H(T) - H(T|Z; \phi)
\end{align*}
where $I(T, Z; \phi)$ is the mutual information between $T$ and $Z$, and $H(T|Z; \phi)$ is the conditional entropy. Marginal entropy $H(T)$ does not depend on $\phi$, thus minimizing KL divergence is equivalent to maximizing the conditional entropy $H(T|Z; \phi) = \mathbb{E}[p_{\phi}(t|z)]$. However, as $p_{\phi}(t|z)$ is intractable we introduce variational distribution $q_{\pi}(t|z)$ defined over the same space to approximate it. 
For any variational distribution $q_\pi(t|z)$ the following holds \cite{aminmaxapproach}.
\begin{align*}
    \max_{\phi}H(T|Z; \phi) = \max_{\phi}\inf_{\pi} \mathbb{E}_{p_{\phi}(t, z)}[-\log q_{\pi}(t|z)]
\end{align*}
We assume the distribution $q_{\pi}(t|z)$ is a normal distribution with a fixed variance. We can estimate the mean of $q_\pi(t|z)$ by a neural network called treatment-prediction network $\pi$. By approximating $p_\phi(z, t)$ with empirical data, we derive the following mean squared adversarial loss term from the above negative log likelihood.
\begin{align*}
    l_{adv} = \max_{\phi} \min_{\pi} \sum_{i=1}^N \frac{1}{N} (t_i - \pi(\phi(x_i)))^2
\end{align*}
Specifically, the treatment-predictor network $\pi$ (The green network in Figure \ref{fig:arch}) is trained to minimize $l_{adv}$ by predicting treatment $t$ from representation $z = \phi(x)$. The encoder network $\phi$ (The blue network in Figure \ref{fig:arch}) is trained to maximize $l_{adv}$ by extracting $z$ in such a way that the assigned treatment $t$ is not distinguishable. Therefore KL divergence can be estimated and minimized using two networks and an adversarial loss. Through alternating optimization with respect to $\phi$ and $\pi$, and assuming that the treatment predictor $\pi$ reaches the optimum in each iteration, the resulting representation $z$ has a desired property: $\mathbb{E}[t|z] = \mathbb{E}[t]$ \cite{wang2020continuously}. This implies that knowing latent representation $z$ does not provide additional information for predicting the expected treatment.
Figure \ref{fig:dist} also demonstrates that the representation learned through KL divergence minimization is less predictive of the treatment value compared to representations obtained from two classes of IPM, HSIC \cite{HSIC} and Wasserstein \cite{Wasserstein}, thereby showing more effective reduction of the shift.
\subsection{Factual Outcome Error Minimization}
In this section, we discuss the minimization of the factual prediction error $\epsilon_f^\ell$. Recall that:
\begin{align*}
    \epsilon_f^\ell &=   \int_{[0, 1]} \int_{\mathcal{X}} \ell_{\phi, h}(x, t)\, p(x, t) \, dx \, dt  \\ 
    &= \int_{[0, 1]} \int_{\mathcal{X}} L(h(\phi(x), t), y)\, p(x, t) \, dx \, dt 
\end{align*}
Here, outcome $y$ is a continuous variable, and we consider $L$ to be the squared loss. By approximating $p(x, t)$ with empirical distribution, we derive the following outcome prediction loss that needs to be minimized with respect to $\phi$ and $h$:
\begin{align*}
     l_{pred} = \min_{\phi, h} \, \sum_{i=1}^N \frac{1}{N} \big(y_i - h(\phi(x_i), t_i)\big)^2
\end{align*}
The encoder network $\phi$ is as defined in the previous section.
The outcome prediction network $h$, however, needs to be particularly designed to maintain the treatment impact on the outcome. 
We aim to obtain an informative embedding for treatment value, and similar to \cite{zhang2022exploring} predict the outcome from the embedding and representation using an attention-based network. \cite{zhang2022exploring} proposed to learn the embedding by a neural network. While neural networks are universal function approximators, it has been shown that they can not extract an expressive embedding from a scalar value due to optimization difficulties \cite{gorishniy2023embeddings}. We construct the treatment embedding applying a set of predefined spline functions to the treatment $t$ shown as $S(t) = [s_1(t), s_2(t), \dots, s_m(t)]$ in Figure 1. Spline functions have been shown to be able to approximate a function in a piece-wise manner \cite{eilers1996flexible}
\newline
\newline
The treatment embedding and the representation are then passed to the cross attention layer (red module in Figure 1) to learn the dependency between treatment and representation. A cross-attention layer has three matrices query $Q$, key $K$, and value $V$, where $Q$ is learned from treatment embedding using $h_q$ parameter and $K$ and $V$ are learned from the representation using $h_k$ and $h_v$ parameters respectively. The output of the cross-attention layer is $ \sigma(\frac{Q^TK}{\sqrt{d_k}})V$ where $d_k$ is the dimension of the $Q$ and $K$ matrices and $\sigma$ denotes the softmax function. We then predict the outcome $\hat{y}$ by a linear layer after the cross attention module. 
\newline
\newline
Unlike ad-hoc architectures VCNet \cite{nie2021vcnet} and DRNet \cite{Schwab_2020}, our outcome prediction network is flexible in terms of the number of spline functions. We can incorporate as many splines as needed without increasing the number of model parameters. This is particularly important in individual effect estimation, because each individual responds differently to a given treatment, and hence different spline functions might be necessary to approximate the treatment-response function for different patients. The proposed architecture can incorporate a large number of spline functions, and the attention layer learns how relevant each spline is for estimating each patient treatment-response function.  
It is also worth mentioning that by setting $h_q$ to the identity, $h_k$ to the unity and parameterizing $h_v$ with a neural network (which are sub-optimal choices) we recover VCNet and DRNet with a cross-attention layer.
\subsection{Adversarial Counterfactual Regression}
\begin{algorithm}[!ht]
\caption{Adversarial CounterFactual Regression}
\begin{algorithmic}[1]
    \State \textbf{Input}: Factual samples {$(x_i, t_i, y_i)_{i=1}^N$}, encoder network with initial parameter $\phi_0$, treatment-predictor network with initial parameters $\pi_0$, hypothesis network with initial parameters $h_0$, batch size $b$, iteration number $T$, inner loop size $M$, trade-off parameter $\gamma$, and the step sizes $\eta_1$ and $\eta_2$.
    \For{$t \leftarrow 0$  \textbf{to} $T-1$}
        \State Sample a mini-batch: $B = \{i_1, i_2,..., i_b\}$:
        \State Encode into latent representation: $z_B = \phi_{t}(x_B)$
        \State Initialize $\omega_0 = \pi_t$
        \For{$m \leftarrow 0$   \textbf{to}  $M-1$}
            \State Compute $l_{adv}$ and update $\omega_{m}$
            \State $\hat{t}_B = \omega_{m}(z_B) \quad l_{adv} = \frac{1}{b} \sum_{i \in B} (t_i - \hat{t_i})^2$
            \State $\omega_{m+1} = \omega_{m} - \eta_2 \gamma \nabla_{\omega} l_{adv}(\omega_m)$
        \EndFor
        \State $\pi_{t+1} = \omega_m$
        \State Compute $l_{pred}$ and update $\phi_t$ and $h_t$:
        \State $\hat{y} = h_t(z_B) \quad l_{pred} = \frac{1}{b} \sum_{i \in B}(y_i - \hat{y_i})^2$
        \State $\phi_{t+1} = \phi_t - \eta_1 \big(\nabla_{\phi} l_{pred}(\phi_t) - \gamma \nabla_{\phi} l_{adv}(\phi_t) \big)$
        \State $h_{t+1} = h_t - \eta_1 \nabla_{h} l_{pred}(h_t)$
    \EndFor
    \State \textbf{Return} $\phi_T$ and $h_T$
\end{algorithmic}
\label{alg:acfr}
\end{algorithm}
In order to minimize the distribution shift, we derived the adversarial loss $l_{adv}$ and introduced the networks encoder $\phi$ and treatment-prediction network $\pi$ to optimize the loss. Similarly, in order to predict factual outcome we derived outcome prediction loss $l_{pred}$ and introduced its associated attention based prediction network $h$ to minimize it. Now, we aim to train the entire network to optimize the following objective function as Proposition 1 suggests:
\begin{align*}
   \mathcal{L}_{ACFR} = \max_{\pi}\min_{\phi, h} l_{pred} - \gamma l_{adv} 
\end{align*}
where $\gamma$ is a tunable parameter. 
Algorithm \ref{alg:acfr} presents pseudo-code for the training of the ACFR network.  At each iteration a batch of samples is given as input to the network (line 3). At the first stage ACFR predicts the treatment using encoder and treatment-prediction networks,
computes $l_{adv}$  and only updates parameters of $\pi$ for $M$
iterations (line 5-10). At the second stage, ACFR predicts
the outcome from the encoded representation of the batch using
outcome prediction network $h$ , computes $l_{pred}$  and updates $h$ and $\phi$
with respect to $l_{pred}$ and $l_{pred} - \gamma \,  l_{adv}$  losses (line 11-14).
Finally, the parameters of the encoder and the prediction networks are returned for the inference phase (line 15).
    
\section{Experiments}
\begin{table*}[ht]
	\centering
	\begin{tabular}[lrr]
		{>{\centering\arraybackslash}p{3cm}>{\centering\arraybackslash}p{1.8cm}>{\centering\arraybackslash}p{1.8cm}>{\centering\arraybackslash}p{1.8cm}>{\centering\arraybackslash}p{1.8cm}}
		\toprule
		 \multicolumn{1}{c}{} & \multicolumn{2}{c}{\textbf{News}} & \multicolumn{2}{c}{\textbf{TCGA}} \\
		Method  & MISE & PE & MISE & PE \\ 
		\midrule
        GPS & $3.21 \pm 0.34$ & $0.39 \pm 0.03$& $6.50 \pm 1.21$ &$ 2.30 \pm 0.27$\\
		 MLP & $2.91 \pm 0.33$ & $0.31 \pm 0.02$ & $4.81 \pm 0.54$& $1.15 \pm 0.23$ \\
   		DRNet-HSIC & $1.59 \pm 0.20$ & $0.21 \pm 0.01$ & $2.03 \pm 0.27$& $1.24\pm 0.23$\\
		 DRNet-Wass &$1.64 \pm 0.21$& $0.21 \pm 0.01$ &$ 2.01 \pm 0.20$& $1.29 \pm 0.21$\\
		 VCNet-HISC & $1.28 \pm 0.10$ & $\mathbf{0.16 \pm 0.01}$& $1.99 \pm 0.11$ &$0.94 \pm 0.14$\\
		 VCNet-Wass &$1.43 \pm 0.11 $& $0.17 \pm 0.01$& $ 1.76 \pm 0.12$& $0.92 \pm 0.14$\\
  		ADMIT-HSIC & $1.25 \pm 0.12$ & $0.18 \pm 0.01$ & $1.81 \pm 0.23$& $ 0.86\pm 0.15$\\
		 ADMIT-Wass &$1.35 \pm 0.20$& $0.18 \pm 0.01$ & $1.67 \pm 0.23$& $0.81 \pm 0.14$\\
   		SCIGAN & $1.21 \pm 0.15$& $0.20 \pm 0.01$& $1.85 \pm 0.14$& $0.97 \pm 0.14$\\
           ACFR w/o attention & $1.58 \pm 0.15$ & $0.19 \pm 0.01$&  ${1.86 \pm 0.21}$& $1.01 \pm 0.15$\\
		 ACFR & $\mathbf{1.12 \pm 0.12}$ & $0.18 \pm 0.01$&  $\mathbf{{1.60 \pm 0.20}}$& $\mathbf{0.76 \pm 0.12}$\\
  \bottomrule
    \end{tabular}
    \caption{Results on News and TCGA datasets for the out-of-sample setting.}\label{table:results}
\end{table*}
\begin{table*}[ht]
	\centering
	\small
	\begin{tabular}
        {>{\centering\arraybackslash}p{1.5cm}
        >{\centering\arraybackslash}p{1.5cm}
        >{\centering\arraybackslash}p{2cm}
        >{\centering\arraybackslash}p{6cm}
        >{\centering\arraybackslash}p{4cm}}
		\toprule
		Dataset & $\#$Samples & $\#$Covariates & Outcome function & Treatment assignment \\
		\midrule
		TCGA & $9659$ & $4000$ & $y = 10 (v_1^Tx + 12v_2^Txt -12 v_3^Txt^2)$ & $t = \text{Beta}(\alpha, \beta)$ \\
		\cmidrule{1-4}
		News & $5000$ & $3477$ & $y = 10(v_1^Tx + \sin( \frac{v_2^Tx}{v_3^Tx}\pi t))$ & $\beta = \frac{2(\alpha - 1)v_2^Tx}{v_3^Tx} + 2 - \alpha$ \\
		\bottomrule
	\end{tabular}
	\caption{Datasets and data generating functions.
        \label{data}}
\end{table*}
Treatment effect estimation methods have to be evaluated for predicting potential outcomes including counterfactuals which are unavailable in real-world observational datasets. Therefore, synthetic or semi-synthetic datasets are commonly used since their treatment assignment mechanism and outcome function are known and hence counterfactual outcomes can be generated. Note that this does not change the fact that only factual outcomes are accessible during training. In this section, we present our experimental results.
The code for synthetic data generation and implementation of the methods can be found at here: https://github.com/amirrezakazemi/acfr
\subsection{Setup}
\subsubsection{Semi-synthetic data generation}
We used TCGA \cite{6c2aadb632ec445491a8f95490e23836} and News \cite{pmlr-v48-johansson16} semi-synthetic datasets. TCGA dataset consists of gene expression measurements of the 4000 most variable genes for 9659 cancer patients. The News dataset which was introduced as a benchmark in \cite{pmlr-v48-johansson16} consists of 3477 word counts for 5000 randomly sampled news items from the NY times corpus. For each dataset, we first normalized each covariate and then scaled every sample to have a norm $1$. We then split the datasets with $68/12/20$ ratio into training, validation, and test sets. We followed treatment and outcome generating process of \cite{bica2020estimating}, summarized in Table 1. The parameter $\alpha$ in treatment function determines the treatment-selection bias level ($\alpha$ is set 2 in all experiments unless otherwise stated), and $v_1$, $v_2$ and $v_3$ are  vectors whose entries are sampled from the normal distribution $\mathcal{N}(0, 1)$, and then are normalized. Using the functions in Table 1, we assigned the treatment and factual outcome for all samples in the training and validation sets. All methods are then trained on the training set, and the validation set has been used for hyperparameter selection. Same as \cite{bica2020estimating}, potential outcomes for a unit are generated using the outcome function given the unit's covariates and $65$ grids in the range $[0, 1]$ as an approximation of the treatment range. 
\subsubsection{Baselines}
DRNet \cite{Schwab_2020} and VCNet \cite{nie2021vcnet} are state-of-the-art neural networks for continuous treatment estimation. Following \cite{bellot2022generalization}, we use the following versions of these two methods as the main baselines. HSIC \cite{HSIC} is a version using the Hilbert-Schmidt independence criterion to minimize the distribution shift while Wass  \cite{Wasserstein} is a version that uses the Wasserstein distance for that purpose. We consider SCIGAN \cite{bica2020estimating} as state-of-the-art generative method for continuous treatments, and also compare against ADMIT \cite{wang2022generalization} network with their proposed algorithm to estimate IPM distances. 
Finally we include Generalized Proposensity Score (GPS) and a MLP network as baselines. The MLP network consists of two fully connected layers without any attempt to reduce distribution shift.
\subsubsection{Metrics}
Having $\mu(x, t)$ as the ground-truth outcome of the unit with covariate $x$ under treatment $t$ and $f(x, t)$ as the predicted outcome, we report the performance of methods in terms of the two following metrics defined in \cite{Schwab_2020}. The Mean Integrated Squared Error (MISE) is the squared error of the predicted outcome averaged over all treatment values and all units. The Policy Error (PE) measures the average squared error of estimated optimal treatment, where $t_{i}^*$ and $\hat{t}_{i}^*$ denote ground-truth and predicted best treatments respectively.
\begin{align*}
    &\text{MISE} = \frac{1}{N}\Sigma_{i=1}^N \int_{0}^1 [\mu(x_i, t) - f(x_i, t)]^2 dt \\
    &\text{PE} = \frac{1}{N} \Sigma_{i=1}^N [\mu(x_i, t_{i}^*) - \mu(x_i, \hat{t}_{i}^*)]^2 \\
    & \quad \
\end{align*}
\subsection{Results}
We performed two sets of experiments for potential outcome prediction, called out-of-sample prediction and within-sample prediction. The out-of-sample experiment shows the ability of models in predicting the potential outcomes for units in the held-out test set, and the within-sample experiment shows the ability for units in the training set.
\subsubsection{Prediction error}
For all methods, we reported the mean and the standard deviation of MISE and PE in the format of mean$\pm$std over 20 realizations of each dataset.
Table 1 shows that on TCGA dataset, ACFR outperformed the contenders in both metrics, and on News dataset, ACFR achieved the best and second best result in terms of MISE and PE metrics respectively.
We can also see the substantial gain of cross-attention layer in the performance of ACFR by comparing it with ACFR w/o attention, demonstrating the effectiveness of proposed outcome prediction network.
Comparably, DRNet and VCNet methods have more parameters while ACFR and ADMIT methods are more time-consuming because of their corresponding inner loop to minimize the distribution shift. You can find the details of our implementation  and the results in the within-sample experiment in Appendix B.
\subsubsection{Treatment-selection bias robustness}
We also investigate the robustness of 4 methods (ACFR, VCNet-HSIC, DRNet-HSIC, and ADMIT-HSIC) against varying level of treatment-selection bias. As mentioned earlier, the $\alpha$ parameter of Beta distribution in the treatment generating function controls the amount of bias. As $\alpha$ increases the treatment-selection bias and covariate shift of the observational dataset increase and consequently, we expect the error of methods to increase as well. 
\begin{figure}[ht] 
    \centering
    \includegraphics[width=0.5\textwidth, height=0.32\textheight]{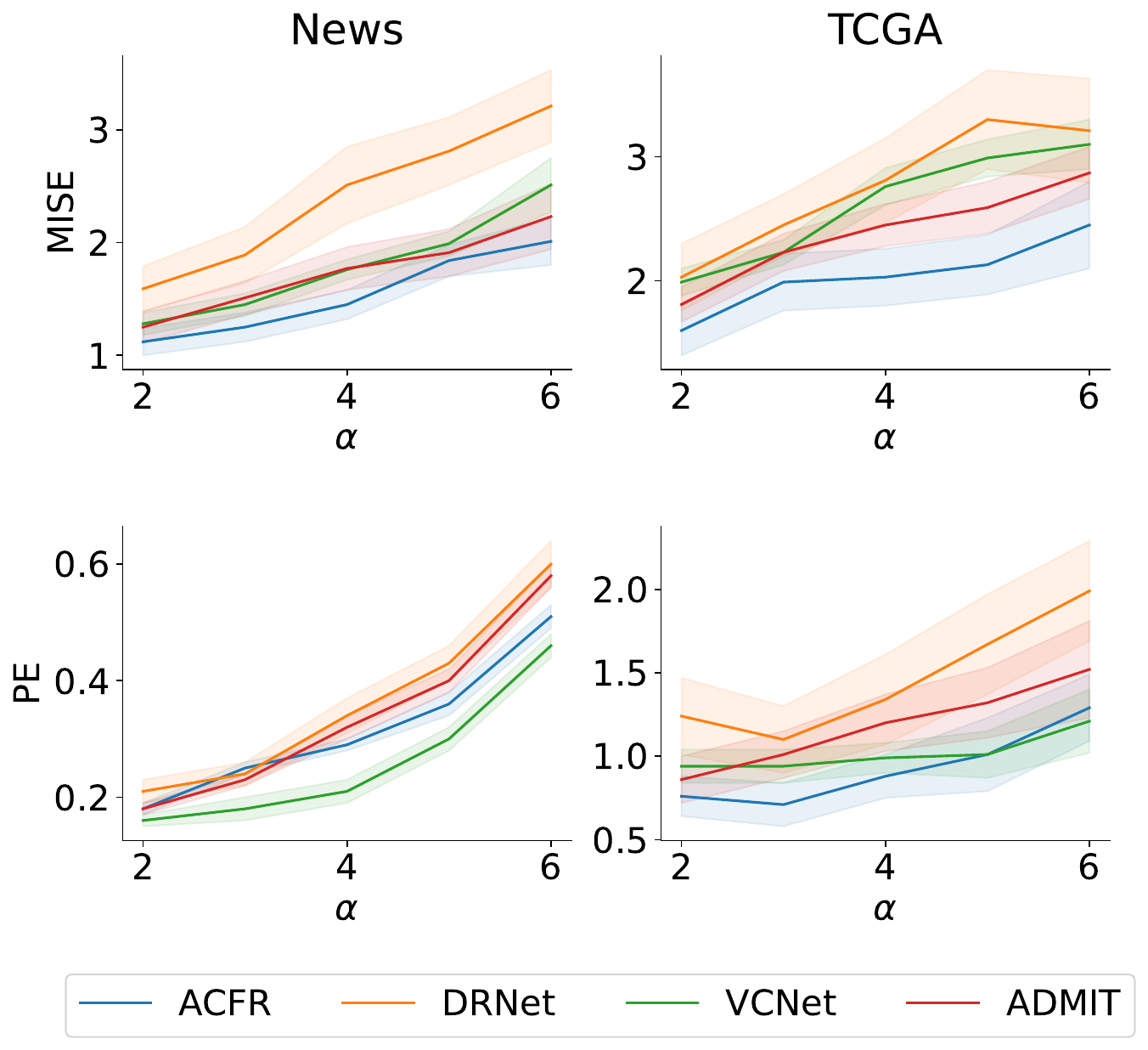} 
    \caption{\textbf{Robustness of ACFR against varying level of treatment-selection bias} determined by $\alpha$ parameter of treatment assignment distribution. ACFR demonstrates a robust performance in terms of MISE and PE compared to baselines.}
    \label{fig:sel}
\end{figure}
As shown in Figure , ACFR performs consistently and has a notable gap with the contenders at the strong bias level ($\alpha$ = 6) in terms of MISE for the out-of-sample setting.

\section{Related Work}
\subsubsection{Continuous Treatment Effect Estimation}
These methods can be categorized into those estimating the average effect and those estimating the individual effect. In the first category, \cite{hirano2004propensity} proposed the generalized propensity score (GPS) that generalizes the notion of propensity score to continuous treatments. \cite{wu2021matching, fong2018covariate} proposed approaches to matching and covariate balancing, respectively, according to the weights learned using GPS.  \cite{nie2021vcnet} proposed the Varying Coefficient network (VCNet) which extracts a  representation sufficient for GPS prediction and predicts the outcome using a network where the treatment value influences the outcome indirectly through parameters instead of being given directly as input.
\cite{bahadori2022endtoend} proposes an entropy balancing method to learn more stable weights compared to GPS based weights. Our approach is fundamentally different since sufficiency of (generalized) propensity score theorem holds only for the average effect and we aim to estimate the individual effect.
\newline
\newline
In the individual effect category, methods are mostly based on learning a balanced representation. DRNet \cite{Schwab_2020} discretized the treatment range into intervals, minimized the pair-wise shift between the populations fall within these intervals in the representation, and proposed a hierarchical multi head network to predict outcomes for the intervals.  \cite{bellot2022generalization} demonstrated that minimizing IPM between $P(Z, T)$ and $P(Z)P(T)$ coupled with factual outcome error leads to an upper bound of the counterfactual error. Nonetheless, in practice they minimized the IPM sample-wise since $P(Z)$ is unknown. Similarly, \cite{wang2022generalization} demonstrated that discretizing the treatment range and minimizing the maximum pair-wise IPM bounds the counterfactual error. Our method is different as we learn the balanced representation by minimizing the KL divergence parameterically.

\subsubsection{Adversarial Balanced Representation}
Learning a balanced (invariant) representation using an adversarial discriminator has been studied in the transfer learning literature to align source domain(s) to target domain(s) \cite{ganin2016domainadversarial, tzeng2017adversarial, wang2020continuously}. Similarly, in causal inference \cite{Du_2021, NEURIPS2020_e7c573c1} aimed to balance the distributions of two treatment groups adversarially. \cite{CRN} extended the approach to the multiple time-varying treatment setting. However, the existing methods consider only scenarios with a finite number of treatment options and do not provide theoretical guarantees of their generalization capability.

\section{Conclusion}
This paper has investigated the problem of continuous treatment effect estimation and introduced ACFR (Adversarial Counter-Factual Regression) method for predicting potential outcomes. We proved a new bound of the counterfactual error using the KL divergence instead of an IPM distance, which has the benefit that the KL divergence can be estimated parametrically and results in a more reliable bound. Based on the error bound, ACFR uses an adversarial neural network architecture to minimize the KL divergence of the representations and a cross-attention network to minimize the factual prediction error.
It is worth mentioning that ACFR is not restricted to continuous treatments, and in future work we plan to extend and evaluate ACFR framework for structured and time series treatments. Nonetheless, we note that ACFR, similar to many treatment effect estimation methods, relies on strong ignorability assumption, which is not necessarily hold in real-world applications.

\section{Acknowledgement}
We would like to thank Oliver Schulte and Sharan Vaswani for feedback on the paper. This research was  supported by the Natural Sciences and Engineering Research Council of Canada (NSERC) Discovery Grant. 
\bibliography{aaai24}

\begin{thebibliography}{42}
\providecommand{\natexlab}[1]{#1}

\bibitem[{Alaa and van~der Schaar(2017)}]{alaa2017bayesian}
Alaa, A.~M.; and van~der Schaar, M. 2017.
\newblock Bayesian Inference of Individualized Treatment Effects using
  Multi-task Gaussian Processes.
\newblock arXiv:1704.02801.

\bibitem[{Bahadori, Tchetgen, and Heckerman(2022)}]{bahadori2022endtoend}
Bahadori, M.~T.; Tchetgen, E.~T.; and Heckerman, D.~E. 2022.
\newblock End-to-End Balancing for Causal Continuous Treatment-Effect
  Estimation.
\newblock arXiv:2107.13068.

\bibitem[{Bareinboim and Pearl(2012)}]{bareinboim2012controlling}
Bareinboim, E.; and Pearl, J. 2012.
\newblock Controlling selection bias in causal inference.
\newblock In \emph{Artificial Intelligence and Statistics}, 100--108. PMLR.

\bibitem[{Bareinboim, Tian, and Pearl(2014)}]{bareinboim2022recovering}
Bareinboim, E.; Tian, J.; and Pearl, J. 2014.
\newblock Recovering from Selection Bias in Causal and Statistical Inference.
\newblock \emph{Proceedings of the AAAI Conference on Artificial Intelligence},
  28(1).

\bibitem[{Bellot, Dhir, and Prando(2022)}]{bellot2022generalization}
Bellot, A.; Dhir, A.; and Prando, G. 2022.
\newblock Generalization bounds and algorithms for estimating conditional
  average treatment effect of dosage.
\newblock arXiv:2205.14692.

\bibitem[{Berrevoets et~al.(2020)Berrevoets, Jordon, Bica, gimson, and van~der
  Schaar}]{NEURIPS2020_e7c573c1}
Berrevoets, J.; Jordon, J.; Bica, I.; gimson, a.; and van~der Schaar, M. 2020.
\newblock OrganITE: Optimal transplant donor organ offering using an individual
  treatment effect.
\newblock In Larochelle, H.; Ranzato, M.; Hadsell, R.; Balcan, M.; and Lin, H.,
  eds., \emph{Advances in Neural Information Processing Systems}, volume~33,
  20037--20050. Curran Associates, Inc.

\bibitem[{Bica et~al.(2020)Bica, Alaa, Jordon, and van~der Schaar}]{CRN}
Bica, I.; Alaa, A.~M.; Jordon, J.; and van~der Schaar, M. 2020.
\newblock Estimating Counterfactual Treatment Outcomes over Time Through
  Adversarially Balanced Representations.
\newblock arXiv:2002.04083.

\bibitem[{Bica, Jordon, and van~der Schaar(2020)}]{bica2020estimating}
Bica, I.; Jordon, J.; and van~der Schaar, M. 2020.
\newblock Estimating the Effects of Continuous-valued Interventions using
  Generative Adversarial Networks.
\newblock arXiv:2002.12326.

\bibitem[{Chu et~al.(2023)Chu, Huang, Li, Chu, and Li}]{chu2023causal}
Chu, Z.; Huang, J.; Li, R.; Chu, W.; and Li, S. 2023.
\newblock Causal Effect Estimation: Recent Advances, Challenges, and
  Opportunities.
\newblock arXiv:2302.00848.

\bibitem[{Du et~al.(2021)Du, Sun, Duivesteijn, Nikolaev, and
  Pechenizkiy}]{Du_2021}
Du, X.; Sun, L.; Duivesteijn, W.; Nikolaev, A.; and Pechenizkiy, M. 2021.
\newblock Adversarial balancing-based representation learning for causal effect
  inference with observational data.
\newblock \emph{Data Mining and Knowledge Discovery}, 35(4): 1713--1738.

\bibitem[{Eilers and Marx(1996)}]{eilers1996flexible}
Eilers, P.~H.; and Marx, B.~D. 1996.
\newblock Flexible smoothing with B-splines and penalties.
\newblock \emph{Statistical science}, 11(2): 89--121.

\bibitem[{Farnia and Tse(2016)}]{aminmaxapproach}
Farnia, F.; and Tse, D. 2016.
\newblock A Minimax Approach to Supervised Learning.

\bibitem[{Fong, Hazlett, and Imai(2018)}]{fong2018covariate}
Fong, C.; Hazlett, C.; and Imai, K. 2018.
\newblock Covariate balancing propensity score for a continuous treatment:
  Application to the efficacy of political advertisements.
\newblock \emph{The Annals of Applied Statistics}, 12(1): 156--177.

\bibitem[{Ganin et~al.(2016)Ganin, Ustinova, Ajakan, Germain, Larochelle,
  Laviolette, Marchand, and Lempitsky}]{ganin2016domainadversarial}
Ganin, Y.; Ustinova, E.; Ajakan, H.; Germain, P.; Larochelle, H.; Laviolette,
  F.; Marchand, M.; and Lempitsky, V. 2016.
\newblock Domain-Adversarial Training of Neural Networks.
\newblock arXiv:1505.07818.

\bibitem[{Gorishniy, Rubachev, and Babenko(2023)}]{gorishniy2023embeddings}
Gorishniy, Y.; Rubachev, I.; and Babenko, A. 2023.
\newblock On Embeddings for Numerical Features in Tabular Deep Learning.
\newblock arXiv:2203.05556.

\bibitem[{Gretton et~al.(2007)Gretton, Fukumizu, Teo, Song, Sch{\"o}lkopf, and
  Smola}]{HSIC}
Gretton, A.; Fukumizu, K.; Teo, C.~H.; Song, L.; Sch{\"o}lkopf, B.; and Smola,
  A. 2007.
\newblock A Kernel Statistical Test of Independence.
\newblock In \emph{NIPS}.

\bibitem[{Hassanpour and Greiner(2019)}]{hassanpour2019learning}
Hassanpour, N.; and Greiner, R. 2019.
\newblock Learning disentangled representations for counterfactual regression.
\newblock In \emph{International Conference on Learning Representations}.

\bibitem[{Hill(2011)}]{Hill2011BayesianNM}
Hill, J.~L. 2011.
\newblock Bayesian Nonparametric Modeling for Causal Inference.
\newblock \emph{Journal of Computational and Graphical Statistics}, 20: 217 --
  240.

\bibitem[{Hirano and Imbens(2004)}]{hirano2004propensity}
Hirano, K.; and Imbens, G.~W. 2004.
\newblock The propensity score with continuous treatments.
\newblock \emph{Applied Bayesian modeling and causal inference from
  incomplete-data perspectives}, 226164: 73--84.

\bibitem[{Johansson, Shalit, and Sontag(2016)}]{pmlr-v48-johansson16}
Johansson, F.; Shalit, U.; and Sontag, D. 2016.
\newblock Learning Representations for Counterfactual Inference.
\newblock In Balcan, M.~F.; and Weinberger, K.~Q., eds., \emph{Proceedings of
  The 33rd International Conference on Machine Learning}, volume~48 of
  \emph{Proceedings of Machine Learning Research}, 3020--3029. New York, New
  York, USA: PMLR.

\bibitem[{Kent, Steyerberg, and van Klaveren(2018)}]{kent2018personalized}
Kent, D.~M.; Steyerberg, E.; and van Klaveren, D. 2018.
\newblock Personalized evidence based medicine: predictive approaches to
  heterogeneous treatment effects.
\newblock \emph{Bmj}, 363.

\bibitem[{Kobrosly(2020)}]{Kobrosly2020}
Kobrosly, R.~W. 2020.
\newblock causal-curve: A Python Causal Inference Package to Estimate Causal
  Dose-Response Curves.
\newblock \emph{Journal of Open Source Software}, 5(52): 2523.

\bibitem[{Liang(2019)}]{liang2019estimating}
Liang, T. 2019.
\newblock Estimating Certain Integral Probability Metric (IPM) is as Hard as
  Estimating under the IPM.
\newblock arXiv:1911.00730.

\bibitem[{Network et~al.(2013)Network, Weinstein, Collisson, Mills, Shaw,
  Ozenberger, Ellrott, Shmulevich, Sander, Stuart, and
  Vandin}]{6c2aadb632ec445491a8f95490e23836}
Network, C.; Weinstein, J.; Collisson, E.; Mills, G.; Shaw, K.; Ozenberger, B.;
  Ellrott, K.; Shmulevich, I.; Sander, C.; Stuart, J.; and Vandin, F. 2013.
\newblock The Cancer Genome Atlas Pan-Cancer analysis project.
\newblock \emph{Nature Genetics}, 45(10): 1113--1120.

\bibitem[{Nie et~al.(2021)Nie, Ye, Liu, and Nicolae}]{nie2021vcnet}
Nie, L.; Ye, M.; Liu, Q.; and Nicolae, D. 2021.
\newblock Vcnet and functional targeted regularization for learning causal
  effects of continuous treatments.
\newblock \emph{arXiv preprint arXiv:2103.07861}.

\bibitem[{Prosperi et~al.(2020)Prosperi, Guo, Sperrin, Koopman, Min, He, Rich,
  Wang, Buchan, and Bian}]{prosperi2020causal}
Prosperi, M.; Guo, Y.; Sperrin, M.; Koopman, J.~S.; Min, J.~S.; He, X.; Rich,
  S.; Wang, M.; Buchan, I.~E.; and Bian, J. 2020.
\newblock Causal inference and counterfactual prediction in machine learning
  for actionable healthcare.
\newblock \emph{Nature Machine Intelligence}, 2(7): 369--375.

\bibitem[{Schwab et~al.(2020)Schwab, Linhardt, Bauer, Buhmann, and
  Karlen}]{Schwab_2020}
Schwab, P.; Linhardt, L.; Bauer, S.; Buhmann, J.~M.; and Karlen, W. 2020.
\newblock Learning Counterfactual Representations for Estimating Individual
  Dose-Response Curves.
\newblock \emph{Proceedings of the {AAAI} Conference on Artificial
  Intelligence}, 34(04): 5612--5619.

\bibitem[{Shalit, Johansson, and Sontag(2017)}]{shalit2017estimating}
Shalit, U.; Johansson, F.~D.; and Sontag, D. 2017.
\newblock Estimating individual treatment effect: generalization bounds and
  algorithms.
\newblock In \emph{International conference on machine learning}, 3076--3085.
  PMLR.

\bibitem[{Sriperumbudur et~al.(2009)Sriperumbudur, Fukumizu, Gretton,
  Schölkopf, and Lanckriet}]{sriperumbudur2009integral}
Sriperumbudur, B.~K.; Fukumizu, K.; Gretton, A.; Schölkopf, B.; and Lanckriet,
  G. R.~G. 2009.
\newblock On integral probability metrics, $\phi$-divergences and binary
  classification.
\newblock arXiv:0901.2698.

\bibitem[{Tzeng et~al.(2017)Tzeng, Hoffman, Saenko, and
  Darrell}]{tzeng2017adversarial}
Tzeng, E.; Hoffman, J.; Saenko, K.; and Darrell, T. 2017.
\newblock Adversarial Discriminative Domain Adaptation.
\newblock arXiv:1702.05464.

\bibitem[{Varian(2016)}]{doi:10.1073/pnas.1510479113}
Varian, H.~R. 2016.
\newblock Causal inference in economics and marketing.
\newblock \emph{Proceedings of the National Academy of Sciences}, 113(27):
  7310--7315.

\bibitem[{Villani(2008)}]{Wasserstein}
Villani, C. 2008.
\newblock Optimal Transport: Old and New.
\newblock In \emph{Optimal Transport: Old and New}.

\bibitem[{Wang, He, and Katabi(2020)}]{wang2020continuously}
Wang, H.; He, H.; and Katabi, D. 2020.
\newblock Continuously indexed domain adaptation.
\newblock \emph{arXiv preprint arXiv:2007.01807}.

\bibitem[{Wang et~al.(2022{\natexlab{a}})Wang, Lyu, Wu, Wu, and
  Chen}]{NEURIPS2022_390bb66a}
Wang, X.; Lyu, S.; Wu, X.; Wu, T.; and Chen, H. 2022{\natexlab{a}}.
\newblock Generalization Bounds for Estimating Causal Effects of Continuous
  Treatments.
\newblock In Koyejo, S.; Mohamed, S.; Agarwal, A.; Belgrave, D.; Cho, K.; and
  Oh, A., eds., \emph{Advances in Neural Information Processing Systems},
  volume~35, 8605--8617. Curran Associates, Inc.

\bibitem[{Wang et~al.(2022{\natexlab{b}})Wang, Lyu, Wu, Wu, and
  Chen}]{wang2022generalization}
Wang, X.; Lyu, S.; Wu, X.; Wu, T.; and Chen, H. 2022{\natexlab{b}}.
\newblock Generalization Bounds for Estimating Causal Effects of Continuous
  Treatments.
\newblock In Oh, A.~H.; Agarwal, A.; Belgrave, D.; and Cho, K., eds.,
  \emph{Advances in Neural Information Processing Systems}.

\bibitem[{Wang et~al.(2020)Wang, Liang, Charlin, and
  Blei}]{10.1145/3383313.3412225}
Wang, Y.; Liang, D.; Charlin, L.; and Blei, D.~M. 2020.
\newblock Causal Inference for Recommender Systems.
\newblock In \emph{Proceedings of the 14th ACM Conference on Recommender
  Systems}, RecSys '20, 426–431. New York, NY, USA: Association for Computing
  Machinery.
\newblock ISBN 9781450375832.

\bibitem[{Wu et~al.(2022)Wu, Yuan, Kuang, Li, Wu, Zhu, Zhuang, and
  Wu}]{wu2022learning}
Wu, A.; Yuan, J.; Kuang, K.; Li, B.; Wu, R.; Zhu, Q.; Zhuang, Y.; and Wu, F.
  2022.
\newblock Learning decomposed representations for treatment effect estimation.
\newblock \emph{IEEE Transactions on Knowledge and Data Engineering}, 35(5):
  4989--5001.

\bibitem[{Wu et~al.(2021)Wu, Mealli, Kioumourtzoglou, Dominici, and
  Braun}]{wu2021matching}
Wu, X.; Mealli, F.; Kioumourtzoglou, M.-A.; Dominici, F.; and Braun, D. 2021.
\newblock Matching on Generalized Propensity Scores with Continuous Exposures.
\newblock arXiv:1812.06575.

\bibitem[{Yao et~al.(2018)Yao, Li, Li, Huai, Gao, and
  Zhang}]{Yao2018RepresentationLF}
Yao, L.; Li, S.; Li, Y.; Huai, M.; Gao, J.; and Zhang, A. 2018.
\newblock Representation Learning for Treatment Effect Estimation from
  Observational Data.
\newblock In \emph{Neural Information Processing Systems}.

\bibitem[{Zhang, Liu, and Li(2021)}]{zhang2021treatment}
Zhang, W.; Liu, L.; and Li, J. 2021.
\newblock Treatment effect estimation with disentangled latent factors.
\newblock arXiv:2001.10652.

\bibitem[{Zhang, Bellot, and van~der Schaar(2020)}]{zhang2020learning}
Zhang, Y.; Bellot, A.; and van~der Schaar, M. 2020.
\newblock Learning Overlapping Representations for the Estimation of
  Individualized Treatment Effects.
\newblock arXiv:2001.04754.

\bibitem[{Zhang et~al.(2022)Zhang, Zhang, Lipton, Li, and
  Xing}]{zhang2022exploring}
Zhang, Y.-F.; Zhang, H.; Lipton, Z.~C.; Li, L.~E.; and Xing, E.~P. 2022.
\newblock Exploring Transformer Backbones for Heterogeneous Treatment Effect
  Estimation.
\newblock arXiv:2202.01336.

\end{thebibliography}

\newpage
\onecolumn
\appendix
\newcommand{\appendixTitle}{%
\vbox{
    \centering

	\vskip 0.2in
	{\LARGE \bf Appendix}
	\vskip 0.2in
	\hrule height 1pt 
}}
\appendixTitle
\section*{Appendix A: Proofs}
\label{app:a}
\begin{restatable*}[\textbf{Proposition 1 - Counterfactual Generalization Bound}]{thm}{prop1} 
\label{prop1}
\textit{
Given the one-to-one encoder function $\phi : \mathcal{X} \rightarrow \mathcal{Z}$, the outcome prediction function $h : \mathcal{Z} \times [0, 1] \rightarrow \mathcal{Y}$, and the unit-loss function $\ell_{L, h, \phi}(x, t)$ that satisfies Assumption 4,
\[
\varepsilon_{cf}^{\ell} \le \varepsilon_f^{\ell} + C\sqrt{ 2 D_{KL}\bigg(p_{\phi}(z, t) || p_{\phi}(z)p(t)\bigg)}
\]
}
\end{restatable*}
\begin{proof}
 Let $\psi : \mathcal{Z} \rightarrow \mathcal{X}$ be the inverse of $\phi$. Similar to the proof technique of \cite{shalit2017estimating}, the following derivations
shows the result
    \begin{align}
        & \varepsilon_{cf}^{\ell} -  \varepsilon_{f}^{\ell} = \int_{[0, 1]}\int_{\mathcal{X}} \ell_{L, h, \phi}(x, t)\, \big[p(x) p(t) - p(x, t) \big]\, dx dt \\
        &= \int_{[0, 1]}\int_{\mathcal{Z}} \ell_{L, h, \phi}(\psi(z), t)\, \big[p(\psi(z))p(t) - p(\psi(z), t)\big] J_{\psi}J_{\psi}^{-1} d\psi(z) dt \\
        &= \int_{[0, 1]}\int_{\mathcal{Z}} \ell_{L, h, \phi}(\psi(z), t)\, \big[p_\phi(z)p(t) - p_\phi(z, t)\big] \,dz dt \\
        &\le  \int_{[0, 1]}\int_{\mathcal{Z}} C \big| p_\phi(z)p(t) - p_\phi(z, t)\big| \\
        &\le  C \sqrt{2\, \int_{[0, 1]}\int_{\mathcal{Z}} p_{\phi}(z) p(t) \log\bigg(\frac{p_{\phi}(z) p(t)}{p_{\phi}(z, t)}\bigg)} \\
        &=  C \sqrt{2\,D_{KL}\bigg(p_{\phi}(z)p(t) || p_{\phi}(z, t)\bigg)}
    \end{align}
    where the equality (3) holds by the reparameterization $x= \psi(z)$, inequality (4)  holds by Assumption 4 constraining the function $\ell$, and the last two inequalities is by Pinkser's inequality $\int |p - q| =2{TV}_{D}(p, q) = \sqrt{2 D_{KL}(p, q)}$.
\end{proof}
In order to prove Proposition 2, we first prove the following lemma.
\\
\begin{restatable*}[\textbf{Lemma 1 }]{thm}{lem1} 
\label{lem1}
\textit{
Given the one-to-one encoder function $\phi : \mathcal{X} \rightarrow \mathcal{Z}$, the outcome prediction function $h : \mathcal{Z} \times [0, 1] \rightarrow \mathcal{Y}$, and the unit-loss function $\ell_{L, h, \phi}(x, t)$ that satisfies Assumption 4, for any treatment $t$ in the valid range,
\[
\varepsilon_{cf}^{\ell}(t) \le \varepsilon_f^{\ell}(t) + C\sqrt{ 2 D_{KL}\bigg(p_{\phi}(z) \big|\big| p_{\phi}(z|t)\bigg)}
\]
}
\end{restatable*}
\setcounter{equation}{0}
\begin{proof}
Let $\psi : \mathcal{Z} \rightarrow \mathcal{X}$ be the inverse of $\phi$.
    \begin{align}
        & \varepsilon_{cf}^{\ell}(t) -  \varepsilon_{f}^{\ell}(t) = \int_{\mathcal{X}} \ell_{L, h, \phi}(x, t)\, \big[p(x) - p(x|t) \big]\, dx \\
        &= \int_{\mathcal{Z}} \ell_{L, h, \phi}(\psi(z), t)\, \big[p_\phi(z) - p_\phi(z|t)\big] \,dz \\
        &\le \int_{\mathcal{Z}} C \big| p_\phi(z) - p_\phi(z|t)\big| \\
        &\le  C \sqrt{2\, \int_{\mathcal{Z}} p_{\phi}(z) \log\bigg(\frac{p_{\phi}(z)}{p_{\phi}(z|t)}\bigg)} \\
        &=  C \sqrt{2\,D_{KL}\bigg(p_{\phi}(z) || p_{\phi}(z|t)\bigg)}
    \end{align}
\end{proof}
Observe the difference between Lemma 1 and Proposition 1. In Lemma 1 the counterfactual error is restricted to a treatment value, and thus it is bounded by the factual error of that specific treatment, and the resulting shift from it. Using this lemma, we can prove the following.
\newline
\newline
\begin{restatable*}[\textbf{Proposition 2 - Precision of Estimating Heterogeneous Effect Bound}]{thm}{prop2}
\textit{Given the same encoder function $\phi$ and outcome prediction function $h$ as in Proposition 1, and a unit-loss function $\ell_{L, h, \phi}(x, t)$ that satisfies Assumption 4 and its associated $L$ is squared error $||.||^2$,
\begin{align*}
   &\varepsilon_{pehe}(t_1, t_2) \le \varepsilon_f^{\ell}(t_1) \, + \, 
\varepsilon_f^{\ell}(t_2) \,  + \, \\
& C \Bigg[\sqrt{ 2 \, D_{KL}\bigg(p_{\phi}(z) || p_{\phi}(z|t_1)\bigg)} +
\sqrt{2\, D_{KL}\bigg(\, p_{\phi}(z) || p_{\phi}(z|t_2)\bigg)} \, \Bigg]
\end{align*}
}
\end{restatable*}
\setcounter{equation}{0}
\begin{proof}
    \begin{align}
        \varepsilon_{pehe}(t_1, t_2) &= \int_{\mathcal{X}} \big[ \big( \mu(x, t_1) - \mu(x, t_2) \big) - \big( h(\phi(x), t_1) - h(\phi(x), t_2) \big) \big]^2 p(x)dx \\
        & \le \int_{X} \big(\mu(x, t_1) -  h(\phi(x), t_1) \big)^2 p(x) dx + \int_X \big( \mu(x, t_2) -  h(\phi(x), t_2) \big)^2 p(x) dx \\
        &= \varepsilon_{cf}^\ell(t_1) + \varepsilon_{cf}^\ell(t_2) \\
        &\le \varepsilon_f^{\ell}(t_1) \, + \, 
        \varepsilon_f^{\ell}(t_2) \,  + \,
        C \Bigg[\sqrt{ 2 \, D_{KL}\bigg(p_{\phi}(z) || p_{\phi}(z|t_1)\bigg)} +
        \sqrt{2\, D_{KL}\bigg(p_{\phi}(z) || p_{\phi}(z|t_2)\bigg)} \, \Bigg]
    \end{align}
    where the inequality (2) is by triangle inequality, and the last two lines hold by the definition of counterfactual error and Lemma 1 respectively.
\end{proof}

\section*{Appendix B: Additional Experiments}
We discuss the implementation and the set of parameters and hyperparameters for each method.
\subsubsection{Adversarial Counterfactual Regression}
For the representation and treatment-predictor networks, we use feedforward layers. To obtain the treatment embedding, we vary the degree $\in \{2, 3, 4\}$ and the knots $\in \{\{1/3, 2/3\}, \{1/4, 2/4, 3/4\}, \{1/5, 2/5, 3/5, 4/5\}\}$ used to construct the spline functions for treatment. For the attention module in the outcome-prediction network, we also consider a feedforward layer with dimensions $32$ for News and $64$ for the TCGA dataset. We finally vary the dimensions of the last layer $\in \{16, 8\}$ and $\{32, 16\}$ for News and TCGA, respectively.

The trade-off parameter in the objective function, $\gamma \in \{10^{-2}, 10^{-1}, 1, 10\}$, and we vary the number of inner loops, $M \in \{1, 10, 100\}$, for optimizing the treatment predictor.

\subsubsection{IPM Minimization Techniques}
We use two different techniques to minimize the IPM distance.

a) As proposed in \cite{bellot2022generalization}, for each sample in a batch, we consider the distance of the joint distribution of that sample $p(z_i, t_i)$ with the distance of the distribution of all other samples $p(z_j, t_j)$ in that batch. This involves computing the IPM term once for each sample as follows:
\[
\frac{1}{N} \sum_{i=1}^N IPM \Big(\{z_i, t_i\}, \{z_i, t_j\}_{j: j \neq i}\Big)
\]

b) As proposed in \cite{wang2022generalization}, we first divide the treatment range into $l$ equal intervals. For each interval $\Delta_i$, we compute the IPM distance of the distribution corresponding to that interval and the distributions of other intervals. We then minimize the maximum distance between them as follows:
\[
\frac{1}{N} \sum_{i=1}^N \sum_{k=1}^l \max \Big( IPM \big( \Delta_k, \Delta_{j: j\neq k} \big) \Big)
\]

We use two classes of IPM families called Hilbert Schmidt Independence Criterion and Wasserstein distance with the implementations from \cite{bellot2022generalization} and \cite{shalit2017estimating} for these two metrics, respectively.

\subsubsection{Multi-Layer Perceptron}
We construct an MLP network using two feedforward layers, which takes the concatenated covariates and treatment as input and predicts the outcome. Its objective is the mean squared error between the ground truth and the predicted outcome.

\subsubsection{Dose-Response Network + IPM}
The original implementation of DRNet considered the treatment variable as a pair of a medication and a dosage, where the medication is categorical and the dosage is a continuous variable. We adjusted the architecture and algorithm for continuous treatment. To minimize the distribution distance in the representation space, we minimize the IPM distance using the first procedure described above. We also use $5$ distinct regression heads for the samples in $5$ equal intervals: $[0, 0.2], [0.2, 0.4], \dots, [0.8, 1]$. Each regression head and the representation network consist of feedforward layers, and similar to the MLP architecture, the treatment value is given to each regression head as input. The weight of the IPM loss term is $\{10^{-3}, 10^{-2}, 10^{-1}, 1\}$ for News and $\{10^{-2}, 10^{-1}, 1, 10\}$ for the TCGA dataset.

\subsubsection{Varying Coefficient Network + IPM}
We adjusted the implementation of VCNet, which was originally proposed for average treatment effect and learned the representation based on propensity score. The adjusted VCNet has two sub-networks. The representation network consists of feedforward layers, and the outcome prediction network is constructed by involving the treatment value into the network parameters. Specifically, we consider a set of spline functions with degree $2$ and knots $[1/3, 2/3]$ and use $5$ heads, each associated with one spline function. The output of the dynamic network is the linear combination of spline functions where the weights are the output of the regression heads. To minimize the distribution distance in the representation space, we minimize the IPM distance using the first technique explained above. The weight of the IPM loss term is $\{10^{-2}, 10^{-1}, 1\}$ for News and $\{10^{-2}, 10^{-1}, 1, 10\}$ for the TCGA dataset.

\subsubsection{ADMIT + IPM}
The IPM minimization procedure in ADMIT is based on the second technique. ADMIT has three sub-networks: representation network, re-weighting network, and hypothesis network. For the representation network, we use feedforward layers, and for the last two, we use the dynamic network proposed by \cite{nie2021vcnet} as described above, in order to maintain treatment impact. The weight of the IPM loss term is $\{10^{-2}, 10^{-1}, 1\}$ for both datasets. Also, we vary the number of intervals: $\{3, 4, 5\}$.

In the feedforward layers of the above implementations, we vary the number of nodes $\in \{50, 100\}$ and the number of hidden layers $\in \{0, 1\}$. The step sizes are $\in \{10^{-5}, 10^{-4}, 10^{-3}\}$ and the batch size is $\{32, 64\}$ for all methods. Additionally, for the Generalized Propensity Score implementation, we employed the implementation in \cite{Kobrosly2020} for continuous outcomes, and adjusted the implementation of SCIGAN \cite{bica2020estimating} for continuous treatments as well. To select the best set of hyperparameters, we used a Bayesian approach, specifically the Tree-Structured Parzen Estimator in the Optuna package.

\subsubsection{Additional Results}
Similar to the results provided for the out-of-sample experiment, we present the prediction error and selection bias robustness results of the within-sample setting for the methods as follows:
\setcounter{table}{0}
\begin{table}[ht]
	\centering
	\begin{tabular}[lrr]
		{>{\centering\arraybackslash}p{3cm}>{\centering\arraybackslash}p{1.8cm}>{\centering\arraybackslash}p{1.8cm}>{\centering\arraybackslash}p{1.8cm}>{\centering\arraybackslash}p{1.8cm}}
		\toprule
		 \multicolumn{1}{c}{} & \multicolumn{2}{c}{\textbf{News}} & \multicolumn{2}{c}{\textbf{TCGA}} \\
		Method  & MISE & PE & MISE & PE \\ 
		\midrule
        GPS & $3.08 \pm 0.33$ & $0.36 \pm 0.02$&  $6.25 \pm 0.97$ &$ 1.95 \pm 0.29$\\
		 MLP & $2.79 \pm 0.32$ & $0.31 \pm 0.02$ & $4.72 \pm 0.65$& $1.27 \pm 0.23$ \\
   		DRNet-HSIC & $1.32 \pm 0.20$ & $0.20 \pm 0.01$ & $1.91 \pm 0.25$& $1.04\pm 0.19$\\
		 DRNet-Wass &$1.34 \pm 0.21$& $0.19 \pm 0.01$ &$ 1.88 \pm 0.20$& $1.04 \pm 0.21$\\
		 VCNet-HISC & $1.18 \pm 0.11$ & $0.16 \pm 0.01$& $1.59 \pm 0.12$ &$0.87 \pm 0.14$\\
		 VCNet-Wass &$1.23 \pm 0.10 $& $0.13 \pm 0.01$& $\mathbf{1.39 \pm 0.11}$& $0.82 \pm 0.14$\\
  		ADMIT-HSIC & $1.12 \pm 0.12$ & $\mathbf{0.12 \pm 0.01}$ & $1.71 \pm 0.23$& $ 0.76\pm 0.15$\\
		 ADMIT-Wass &$1.20 \pm 0.10$& $0.13 \pm 0.01$ & $1.46 \pm 0.21$& $0.79 \pm 0.13$\\
   		SCIGAN &$1.15 \pm 0.11$& $0.16 \pm 0.01$ & $1.58 \pm 0.21$ & $0.86 \pm 0.10$ \\
           ACFR w/o attention & $1.34 \pm 0.13$ & $0.17 \pm 0.01$&  ${1.66 \pm 0.20}$& $0.92 \pm 0.13$\\
		 ACFR & $\mathbf{0.95 \pm 0.12}$ & $0.15 \pm 0.01$&  ${1.42 \pm 0.22}$& $\mathbf{0.62 \pm 0.11}$\\
  \bottomrule
    \end{tabular}
    \caption{Results on News and TCGA datasets for the within-sample setting.}
    \label{table:results2}
\end{table}

\setcounter{figure}{0}
\begin{figure}[ht] 
    \centering
    \includegraphics[width=0.7\textwidth, height=0.6\textheight]{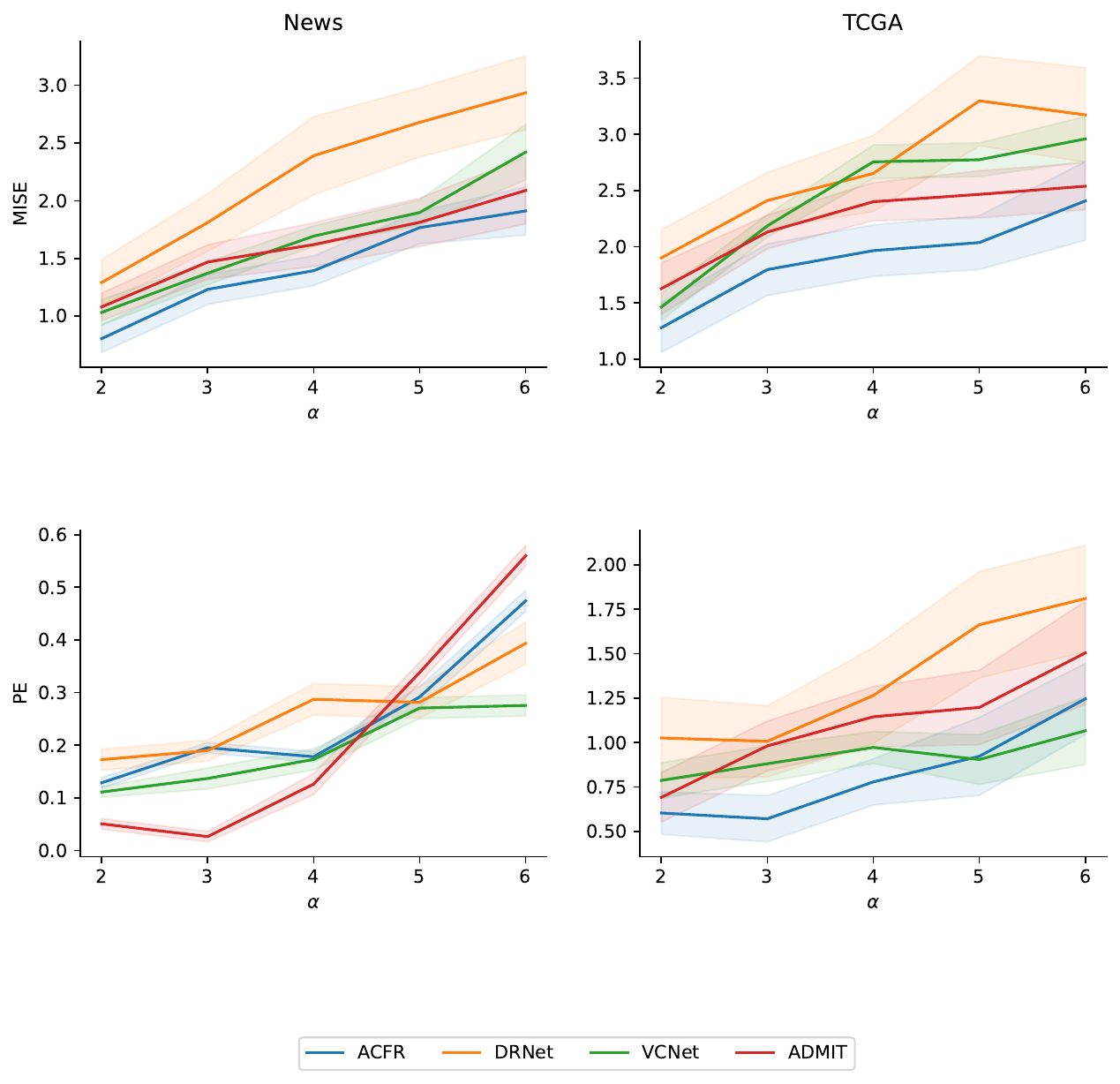} 
    \caption{\textbf{Robustness of ACFR against varying level of selection bias in within-sample setting} determined by $\alpha$ parameter of treatment assignment distribution. ACFR demonstrates a robust performance in terms of MISE and PE compared to contenders.}
    \label{fig:sel2}
\end{figure}

\end{document}